\documentclass[conference]{IEEEtran}
\IEEEoverridecommandlockouts
\usepackage{cite}
\usepackage{algorithmic}
\usepackage{graphicx}
\usepackage{textcomp}
\usepackage{xcolor}
\usepackage{graphics}
\usepackage[switch]{lineno}
\usepackage{epsfig}
\usepackage{caption}
\usepackage{subcaption}
\usepackage{amssymb}
\usepackage{amstext}
\usepackage{amsmath}
\usepackage{enumitem}

\usepackage{multirow}
\usepackage{multicol}
\usepackage{booktabs}
\usepackage{floatrow}
\newfloatcommand{capbtabbox}{table}[][\FBwidth]
\newtheorem{problem}{Problem}
\usepackage{floatrow}
\usepackage{xspace}
\usepackage{color}
\usepackage{xcolor}
\usepackage[most]{tcolorbox}
\usepackage{footmisc}
\usepackage[linesnumbered,ruled,vlined]{algorithm2e}

\usepackage{makecell}
\usepackage{bigstrut}
\usepackage{float}
\usepackage{enumerate}
\newfloat{figtab}{htb}{fgtb}
\makeatletter
  \newcommand\figcaption{\def\@captype{figure}\caption}
  \newcommand\tabcaption{\def\@captype{table}\caption}
\makeatother

\DeclareMathOperator*{\argmax}{arg\,max}
\DeclareMathOperator*{\argmin}{arg\,min}

\definecolor{babyblueeyes}{rgb}{0.19, 0.55, 0.91}

\newcommand{\linecomment}[1]{\textcolor{babyblueeyes}{\textit{$\triangleright$ #1}}}

\newcommand{\nosemic}{\renewcommand{\@endalgocfline}{\relax}}
\newcommand{\dosemic}{\renewcommand{\@endalgocfline}{\algocf@endline}}
\let\oldnl\nl
\newcommand{\nonl}{\renewcommand{\nl}{\let\nl\oldnl}}

\newcommand{\modelset}{\mathcal{M}_{K}^{*}\xspace}

\begin{document}

\title{Data-Free Diversity-Based Ensemble Selection For One-Shot Federated Learning in Machine Learning Model Market}


\author{
    \IEEEauthorblockN{Naibo Wang, Wenjie Feng, Fusheng Liu, Moming Duan, See-Kiong Ng}
    \IEEEauthorblockA{Institute of Data Science, National University of Singapore}
    \IEEEauthorblockA{naibowang@u.nus.edu; wenchiehfeng.us@gmail.com; fusheng@u.nus.edu; moming@nus.edu.sg; seekiong@nus.edu.sg}
}

%

\hyphenation{op-tical net-works semi-conduc-tor down-sampl-ing im-balanced im-balance hetero-geneous FedProx FedAvg FEMNIST}

\newcommand{\methodName}{\texttt{DeDES\xspace\ }}

\maketitle

\begin{abstract}

The emerging availability of trained machine learning models has put forward the novel concept of \textit{Machine Learning Model Market} in which one can harness the collective intelligence of multiple well-trained models to improve the performance of the resultant model through one-shot federated learning  and ensemble learning in a data-free manner.  

However, picking the models available in the market for ensemble learning is time-consuming, as using all the models is not always the best approach. It is thus crucial to have an effective \textit{ensemble selection} strategy that can find a good subset of the base models for the ensemble.  Conventional ensemble selection techniques are not applicable, as we do not have access to the local datasets of the parties in the federated learning setting. In this paper, we present a novel \textit{Data-Free Diversity-Based} method called \methodName to address the ensemble selection problem for models generated by one-shot federated learning in practical applications such as model markets. 
Experiments showed that our method can achieve both better performance and higher efficiency over 5 datasets and 4 different model structures under the different data-partition strategies.

\end{abstract}


\section{Introduction}
\label{sec:introduction}


To address the increasing demands on  data privacy protection while satisfying the growing appetites for more data for machine learning tasks, 
federated learning~\cite{li2020federated} (FL) has become the mainstay for enabling collaborative machine learning on decentralized devices/parties without seeing any of their data.
However, traditional multi-round federated learning training process  has its drawbacks: for $m$ clients and $n$ training rounds, the server can acquire $\mathcal{O}(mn)$ gradients or models, which can possibly reveal a great deal of sensitive information of the clients' local data and violate the privacy protection setting \cite{geiping2020inverting}.

\textit{One-shot federated learning} \cite{guha2019one} has been proposed to further protect the privacy of clients, by only requiring the clients to  send their final well-trained models to the server once. 
In this way, not only the privacy of clients can be better protected,  the communication costs are also significantly decreased. However, the model generated by one-shot federated learning is often less accurate than the model generated by conventional federated learning. As a result, the one-shot federated learning method is unsuitable for  applications such as medical diagnosis, where the model's accuracy is crucial. 

The emerging availability of pre-trained machine learning models for various machine learning tasks has put forward the novel concept of \textit{Machine Learning Model Market} \cite{ijcai2021p205} (beyond model management systems like modelDB \cite{vartak2016modeldb} or huggingface \cite{wolf2019huggingface}) to harness the collective intelligence from multiple well-trained models in a data-free manner. Clients can upload their individual well-trained models to the market server, and the server can select multiple models from its database and perform collective machine learning  (e.g., ensemble learning or model fusion) to enhance the performance of the targeted machine learning task (e.g. image or text classification).

Compared to model fusion, ensemble learning \cite{sagi2018ensemble} is comparatively straightforward and cost-effective  to  harness the power of collective machine intelligence to boost task performance in a data-free manner. For example, a classic ensemble learning method is the \textit{Voting} method by which multiple models will vote together to produce the final classification results. However, selecting all available models from the model market for ensemble learning is not always the most effective strategy. As shown by Zhou et al. \cite{zhou2002ensembling},  \textbf{many could be better than all} when ensembling neural networks. In addition, testing each incoming sample $m$ times, when we have a large number of $m$ models in an ensemble team, can also be  time-consuming and inefficient. As such, we focus on the \textit{ensemble selection} or \textit{ensemble pruning} \cite{caruana2004ensemble} problem, which aims to find a good subset of base models for ensemble from the model market.



A key consideration for ensemble selection is the \textit{model diversity}. Numerous papers have demonstrated that the more diverse the models, the higher the ensemble's performance will have \cite{wu2021boosting,wu2021boosting2}. While numerous model diversity calculation methods have been proposed  to maximize model diversities, they typically require access to the local datasets of the parties which is not possible in the one-shot federated learning setting of model markets.   As such, none of the existing methods can be utilized to calculate model diversity within an ensemble team under the one-shot federated learning setting.



In this work, we propose a novel \textit{\textbf{D}ata-Fre\textbf{e} \textbf{D}iversity-Based} \textbf{E}nsemble \textbf{S}election framework called \textbf{\methodName} for selecting strong ensemble teams for ensemble learning, whose models are sourced from the machine learning model market and trained by one-shot federated learning. We perform a series of studies to show that our presented method is robust, efficient, and successful for various data partitions (especially non-i.i.d data), datasets, and model structures. To the best of our knowledge, this is the first paper to systematically deal with the problem of ensemble selection for one-shot federated learning, which is a valuable application for machine learning model market.


\begin{figure*}[t]
    \centering
    \includegraphics[width=1.0\textwidth]{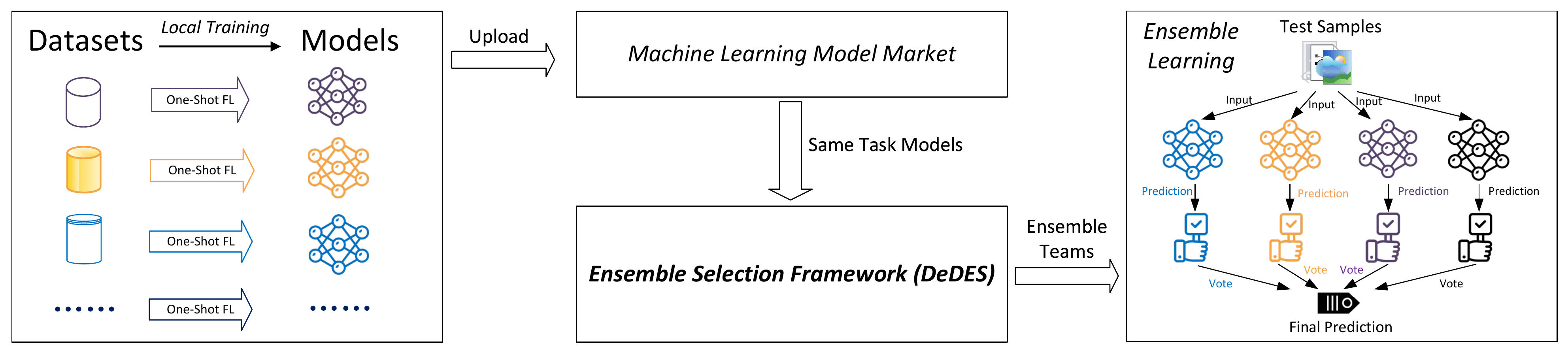}
    \caption{Overview of ensemble learning and ensemble selection process on machine learning model market under one-shot federated learning setting.}
    \label{fig:overview}
\end{figure*}


Fig. \ref{fig:overview} depicts our scenario. Clients will train their models locally by their own dataset until convergence and then upload their models to the model market. To conduct ensemble learning, the server will select, based on our algorithm, a good ensemble team from all models with the same task on the model market. Note that during the whole process, the server has no access to the local datasets of clients at all, which is what we mean by \textit{data-free}. 

The contributions of our paper are as follows:

1. We proposed a formal formulation of the ensemble selection problem  to facilitate a clearer comprehension of the topic;

2. We presented a \textit{Data-Free Diversity-Based} Ensemble Selection framework \methodName for \textit{One-Shot Federated Learning} which can evaluate model diversity and conduct ensemble pruning with no data exposure;

3. We proposed a technique for selecting the representative model inside a cluster to improve the performance of the final ensemble learning; and

4. We conducted a set of comprehensive experiments   to illustrate the efficacy and efficiency of the proposed ensemble selection approach.


Our codes and supplementary material are available online~\footnote{https://anonymous.4open.science/r/DeDesForOSFL/}.


\section{\uppercase{Related Work}}
\label{sec:related_work}

Various federated learning systems \cite{bonawitz2019towards,duan2020self} have been proposed to assist various parties in cooperatively training a global model without disclosing their data. In particular, one-shot federated learning proposes to train a global model using a single round of server-client communication. \textit{FedKT} \cite{ijcai2021p205}, \textit{Fusion Learning} \cite{kasturi2020fusion}, etc. are all good examples of one-shot federated learning; however, none of them tackle the ensemble selection problem for one-shot federated learning.


Lately, with the popularity of utilizing pre-trained models,  there is emerging interest in \textit{Machine Learning Model Market} \cite{bommasani2021opportunities} as a platform for users to exchange their trained models from others, and to harness the collective intelligence for the targeted machine learning task by combine the models. Note that the model market is a concept differs from previous concepts such as model management systems like \textit{ModelDB} \cite{vartak2016modeldb} or huggingface \cite{wolf2019huggingface} which only include the fundamental model manipulation features of upload, download, and search. Or \textit{TFX} \cite{baylor2017tfx} which aims to deploy production ML pipelines. The goal of model market is to enable collaborative machine learning through utilizing the collective intelligence of multiple machine learning models, using model sharing, model unlearning, model pruning, model compression, model valuation, model recommendation, model ensemble, etc. 


Compared to federated learning, ensemble learning, which seeks to merge multiple weak learners (base models) into strong learner(s), has been a  popular topic for decades. \textit{Voting} \cite{zhou2021ensemble}, \textit{Bagging} \cite{sagi2018ensemble}, \textit{Boosting} \cite{schapire2013explaining}, and \textit{Stacking} \cite{wang2019stacking} are examples of traditional ensemble learning approaches. \textit{Ensemble Selection} is an important concern in ensemble learning.  There are three major  approaches  to select a fixed ensemble team for every incoming test sample: \textit{Search-based} \cite{caruana2004ensemble}, \textit{rank-based} \cite{ma2015several}, and \textit{cluster-based} \cite{maskouni2018auto}. Cluster-based ensemble selection approaches are based on model diversity. Classic model diversity calculation methods include \textit{Binary Disagreement} \cite{kuncheva2003measures}, \textit{Cohen’s Kappa} \cite{mchugh2012interrater}, \textit{Q Statistics} \cite{yule1900vii}, \textit{Generalized Diversity} \cite{partridge1997software} and \textit{Kohavi-Wilpert Variance} \cite{kuncheva2003measures}. All of these methods require access to the local dataset and thus violates the fundamental constraint of federated learning.



\section{Problem Definition}
\label{sec:preliminary}


Assume that there are $m$ different clients as parties  
who want to collaborate in machine learning on a given ML task, e.g., classification or regression.
Let $\mathcal{M} := \{M_1, \ldots, M_m\}$ be the well-trained models with each $M_i$ 
trained on $i$-th client via the one-shot federated learning strategy 
over its private dataset $D_i = \{(x_k, y_k)\}_{k=1}^{n_i}$ with size $n_i$, 
where each data is i.i.d. sampled from an unknown distribution $\mathcal{D}$.
 $\mathcal{M} $ will then be uploaded to the central server of the machine learning model market. 
Our ensemble selection problem can be formulated as:

\begin{problem}
\label{prob:model_select}
\textbf{Given:} the model set $\mathcal{M}$ and a relative small constant $K < m$,
\textbf{find} the optimal subset $\modelset$ of $\mathcal{M}$ such that
\begin{equation}
    \label{eq:esdef}
       \mathcal{M}_{K}^{*} = \argmin_{\mathcal{M}_{K} \subseteq \mathcal{M}, |\mathcal{M}_{K}| = K} 
                         \mathbb{E}_{(x, y) \sim \mathcal{D}} \, \ell(y, f_{\mathcal{M}_{K}}(x))),
\end{equation}
where $f_{\mathcal{M}_{K}}(\cdot)$ is the prediction function based on $\mathcal{M}_{K}$ and 
$\ell$ is the loss function.
\end{problem}

Under the ensemble learning setting, $f_{\mathcal{M}_{K}}$ is the aggregation function
to combine the prediction of $M_i \in \mathcal{M}_{K}$ for the final prediction $\hat{y} = f_{\mathcal{M}_{K}}(x)$;
it can be weighted average for regression, or weighted voting-based (e.g., majority or plurality voting) for classification.
Under the model fusion setting, $f_{\mathcal{M}_{K}}$ is the prediction of the fusion model based on all elements in $\mathcal{M}_{K}$.

We focus on the classification task in the following sections 
and we adopt the weighted voting strategy based on the size of local clients' datasets for ensemble learning. 
Thus, for a $C$-class classification (i.e., the label set is $\{1, \ldots, C\}$) task, with $\mathbb{I}(\cdot)$ as the indicator function, the prediction $\hat{y}$ of the input $x$ is given by
\begin{equation}
    \label{eq:weightedVoting}
     \hat{y} := \underset{ c \in \{1, \ldots, C\}} \argmax
    \sum\limits_{j=1}^{K}
    \frac{n_i}{\sum_{k=1}^{K}n_k}\mathbb{I}\left(M_{j}(x) = c\right),
\end{equation}

\section{Proposed Framework: \methodName}
\label{sec:es}

We present our proposed ensemble selection framework, \methodName, to solve Problem~\ref{prob:model_select}
without accessing to any dataset from the local clients.
Algorithm~\ref{alg:method_algframe} summarizes the structure of \texttt{DeDES}.  
(An illustrative view is given in supplementary.)

\begin{algorithm}[t]
    \caption{\methodName framework}
    \label{alg:method_algframe}
        \KwIn{model set $\mathcal{M}$, 
              model training-set sizes $\mathcal{N} = \{n_i\}_{i=1}^{m}$, 
              model scores $\mathcal{S} = \{s_i\}_{i=1}^{m}$,
              and target model-set size $K$.}
        \KwOut{Optimal model subset (Ensemble Team) $\modelset$.}
        
        $\modelset \leftarrow \emptyset$
        
        \nonl \linecomment{1. Model filtering: select high-quality candidates by filtering out outliers.}
        
        $\mathcal{O} \leftarrow \mathtt{OutlierFilter}(\mathcal{M}, \mathcal{S})$   \hfill \textcolor{babyblueeyes}{\linecomment{$\mathcal{O}$: outlier models set}}
        
        $\mathcal{N} = \{n_i ~|~M_i \notin \mathcal{O}, \forall M_i \in \mathcal{M}\}$; \,
        $\mathcal{S} = \{s_i ~|~M_i \notin \mathcal{O}, \forall M_i \in \mathcal{M}\}$; \,
        $\mathcal{M} = \mathcal{M} \setminus \mathcal{O}$;
        
        \nonl \linecomment{2. Model representation: get model's feature representation.}
        
        $\mathcal{R}_{\mathcal{M}} = \{ R_i = \mathtt{Represention}(M_i)~|~ \forall M_i \in \mathcal{M} \}$
        
        \nonl \linecomment{3. Model clustering: get $K$-size model clusters for diversity selection.}
        
        $\mathcal{C}_{\mathcal{M}} = \mathtt{Clustering}(\mathcal{R}_{\mathcal{M}}, K)$
        
        \nonl \linecomment{4. Representative model selection: choose the `best' model in each cluster.}

        \For{$\mathcal{C} \in \mathcal{C}_{\mathcal{M}}$}{
            $\mathcal{N}_{\mathcal{C}} = \{n_i ~|~ \forall M_i \in \mathcal{C} \cap \mathcal{M} \}$
            
            $n_{\max}^{\mathcal{C}} \leftarrow \max(\mathcal{N}_{\mathcal{C}}); \, n_{\mathrm{med}}^{\mathcal{C}} \leftarrow \mathrm{median}( \mathcal{N}_{\mathcal{C}})$
            
            \nonl  \textcolor{babyblueeyes}{\linecomment{$\tau$: user predefined threshold, e.g., $\tau = 0.3$.}} 
            
            \If{$n_{\mathrm{med}}^{\mathcal{C}} ~/~ n_{\max}^{\mathcal{C}}  < \tau$}{
                 $k = \argmax_{j} \{n_j ~|~ M_j \in \mathcal{C} \cap \mathcal{M} \}$
            }
            \Else{
                $k = \argmax_{j} { \{s_j ~|~ M_j \in \mathcal{C} \cap \mathcal{M} \} }$
            }
            
            $\modelset \leftarrow  \modelset \cup \{\mathcal{C}_{k}\}$
            \hfill \textcolor{babyblueeyes}{\linecomment{$\mathcal{C}_{k}$: the $k$-th element of the cluster $\mathcal{C}$}}
        }
        
        \Return{$\modelset$}
\end{algorithm}

Considering the performance and efficiency of $\modelset$, 
it is necessary to choose a small $K$ and keep the diversity and high-quality among selected elements/models.
\methodName achieves such goal via different components,
including \textit{model filtering}, \textit{model representation}, \textit{model clustering}, 
and \textit{representative model selection}, which are explained in detail as follows.

\paragraph{Model filtering:} 
Being from multiple parties in FL, the performance of those various models can vary significantly
and are out-of-control to the central server.
The inferior model may result from different reasons,
including low-quality training data, e.g., being unreliable or contaminated, and with much noise,
trained with inappropriate parameters, etc.
Therefore, it is necessary to filter out such outlier models to eliminate the effect of the noises 
and help to select high-quality models efficiently.
In Alg.~\ref{alg:method_algframe}, we use the $\mathtt{OutlierFilter}$ to obtain the outlier models $\mathcal{O}$ 
based on the model scores $\mathcal{S}$ provided from each party, 
which can be the local validation accuracy or prediction confidence.
$\mathtt{OutlierFilter}$ can be any score-based unsupervised outlier detection methods~\cite{zhao2019pyod}, 
we used a variation of the commonly-used box-plot in our experiment (refer to the supplementary).


\paragraph{Model representation:}

Given the model structure and its parameters, generating effective and suitable representation for the models 
is crucial to measure their properties, like similarity and diversity.
Intuitively, we can use all or partial (some layers) of the parameters to represent the model.
Considering that all models in $\mathcal{M}$ are of the same type,
we choose to use the parameters of the last layer of the model, 
which contain individualized and sufficient information about the model behavior (especially for the classifier) 
and data manifold/space for local training.
Besides, to distill compact information and suppress noise for the representation, especially for big models like Resnet-101,
dimension reduction (DR) is also applied for the representations;
all unsupervised approaches can be adopted here, including the classical PCA, Kernel-PCA, and so on.

In the Alg.~\ref{alg:method_algframe}, we obtain the presentation $R_i$ for the model $M_i$ via the function $\mathtt{Represention}$ in Line~4, 
which extracts the parameters of the last layer of $M_i$ to be a flatten vector and conducts dimension reduction for the vector after normalization. The target dimension for DR is set to be $|\mathcal{M}|$ by default.

\paragraph{Model clustering:} 
To guarantee the diversity in $\modelset$ as we mentioned before,
we can utilize the clustering method to identify the similarity of different models, 
where models with similar properties are grouped into the same cluster
and different clusters are as different as possible.
We can use the  traditional clustering approach here, 
such as K-Means, Hierarchical Clustering, and Spectral Clustering, etc.
and set the target number of clusters as $K$.
This process is denoted by $\mathtt{Clustering}$ in Alg.~\ref{alg:method_algframe}, 
which leads to $\mathcal{C}_{\mathcal{M}}$ as the resultant clusters.

\paragraph{Representative model selection:} 
To choose exactly $K$ models with high performance,
we elaborately select the representative element in each cluster while keeping the diversity. 
Among the models in each cluster, we can intuitively select 
the model with either the highest model score $s_i \in \mathcal{S}$ (provided by the individual party) or 
the largest training dataset (leading to a better-trained model).

Therefore, as the Line 6-13 in Alg.~\ref{alg:method_algframe} shows, 
we design a heuristic select strategy to make full use of these two ways, which can choose a better one than any of the fixed way as the experiment results proved. 
That is, if the amount of training data for the models inside the cluster is balanced (measured by the ratio between the median size and the maximum size), 
the model with the highest model score is chosen,
otherwise, the one with the largest training data is chosen.





\paragraph{Inference:}
After obtaining the optimal $\modelset$  with Algorithm~\ref{alg:method_algframe},
we will conduct ensemble learning with the weighted voting as Eq.~\eqref{eq:weightedVoting}. 
Note that in the whole process of \methodName, 
we successfully select the ensemble team $\modelset$ based on the \textit{model diversity} 
without accessing to any of \textit{local private data} of these parties.

\begin{figure*}[t]
    \centering
    \subcaptionbox{homo \label{subfig:a}}[.23\linewidth]
    {%
        \includegraphics[width =\linewidth]{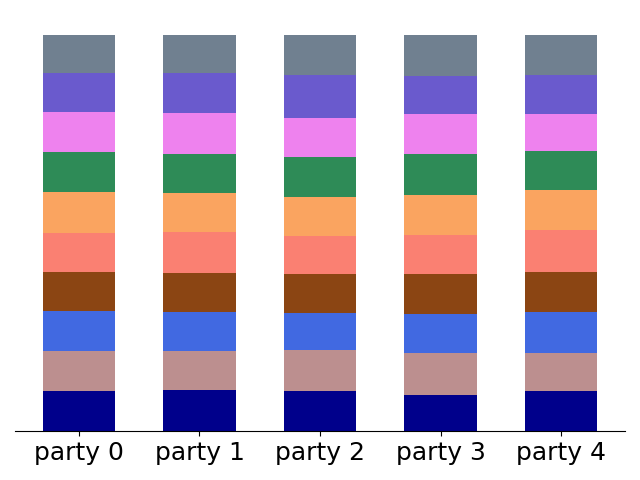}}\quad
    \subcaptionbox{iid-dq \label{subfig:b}}[.23\linewidth]
    {%
        \includegraphics[width =\linewidth]{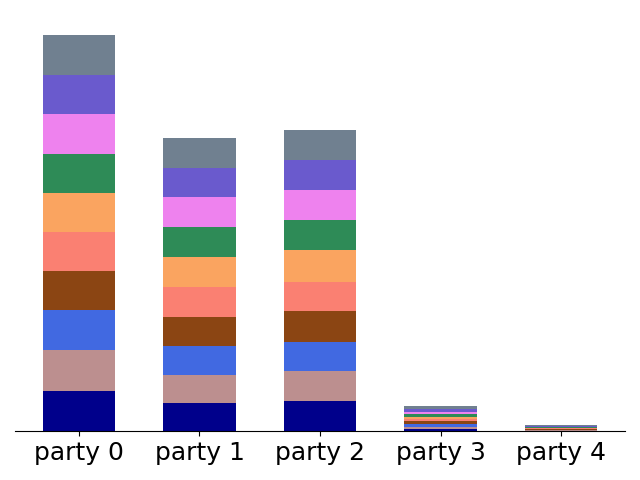}}
    \subcaptionbox{noniid-lds \label{subfig:c}}[.23\linewidth]
    {%
        \includegraphics[width =\linewidth]{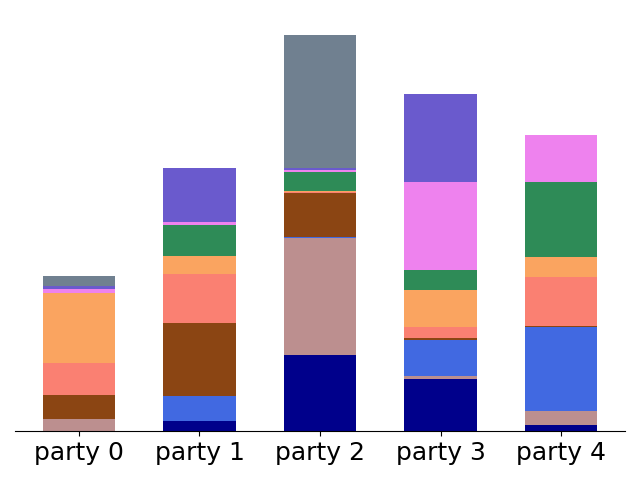}}\quad
    \subcaptionbox{noniid-l$4$ \label{subfig:d}}[.23\linewidth]
    {%
        \includegraphics[width =\linewidth]{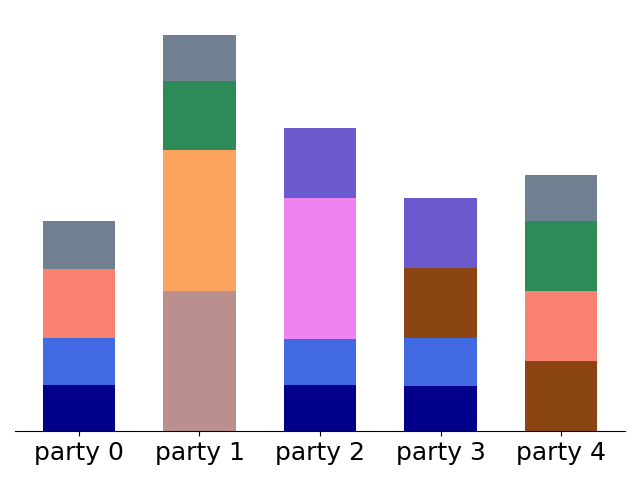}}
    \caption{Example distribution of four dataset partition strategies for the \textit{CIFAR10} dataset with party number $m=5$. Every color bar shows a different class and the height of the bar represents the number of samples of that class.}
    \label{fig:dataset}
\end{figure*}

\section{\uppercase{Experiments}}
\label{sec:exp_setup}

\subsection{Experiment Setup}

To simulate the real scenarios in federated learning as~\cite{li2022federated} 
and comprehensively evaluate \texttt{DeDES},
we designed four types of dataset-partition strategies as follows, which lead to 
different local data distribution to train diverse models $M_i$s. 
\begin{itemize}
    \item Homogeneous~(\emph{homo}): the amount of samples and the data distribution keep the same for all parties;
    \item IID but different quantity~(\emph{iid-dq}): the training data of each party follows the same distribution, but the amount of data is different; 
    \item Skewed data distribution~(\emph{noniid-lds}): the training data of each party follows different distributions, especially for the label distribution;
    \item Non-IID with $k$ ($ < C$) classes~(\emph{noniid-l`$k$'}): the training data of each party only contains $k$ of $C$ classes, which is an extreme Non-IID setting.
\end{itemize}

We used 5 image datasets and 4 types of neural network models (i.e., VGG-5, ResNet-50, DenseNet-121, and Deep Later Aggregation) in our experiments.
Table~\ref{tab:expconfig} lists the detailed information about the datasets and configurations.
We partition all datasets into different groups based on the above strategies and train the model for each client.
Fig.~\ref{fig:dataset} shows an example for the data distribution under the different partition strategies 
for CIFAR10 with 5 parties.

\begin{table*}[tbp]\scriptsize
    \centering
    \caption{Details of experiment configurations}
    \begin{tabular}{c|c|c|c | c|c}
        \hline
        Dataset & \makecell{$C$} & Size ($\sum_{i}n_i$)  & \makecell{$k$ in \\ noniid-l$k$} & Model & \makecell{$m$} \\
        \hline
        EMNIST Digits & 10    & 280,000 & 3     & \multicolumn{1}{c|}{\multirow{2}[6]{*}{\makecell{VGG-5 (Spinal FC),\\ResNet-50}}} & \multirow{2}[6]{*}{100, 200, 400} \\
        \cline{1-4}EMNIST Letters & 26    & 145,600 & 8     &       &        \\
        \cline{1-4}EMNIST Balanced & 47    & 131,600 & 18    &       &        \\
        \hline
        \multirow{3}[2]{*}{CIFAR10} & \multirow{3}[2]{*}{10} & \multirow{3}[2]{*}{60,000} & \multirow{3}[2]{*}{4} & \multicolumn{1}{c|}{\multirow{3}[2]{*}{\makecell{ResNet-50,\\DenseNet-121}}} & \multirow{3}[2]{*}{50, 100, 200} \\
              &       &       &       &       &        \\
              &       &       &       &       &        \\
        \hline
        \multirow{3}[2]{*}{CIFAR100} & \multirow{3}[2]{*}{100} & \multirow{3}[2]{*}{60,000} & \multirow{3}[2]{*}{45} & \multicolumn{1}{c|}{\multirow{3}[2]{*}{\makecell{ResNet-50,\\Deep Layer Aggregation}}} & \multirow{3}[2]{*}{5, 10, 20} \\
              &       &       &       &       &        \\
               &       &       &       &       &        \\
        \hline
        \end{tabular}%
        
    \label{tab:expconfig}%
\end{table*}%

The detailed configuration information of \methodName are elaborated in the supplementary, 
including the learning rate, model representation strategy, clustering method for different data partitions, etc.

\normalsize






\subsection{Baseline Strategies} 
\label{subsec:baseline}

For the model ensemble learning under our problem setting, 
we follow the designs in~\cite{guha2019one} and 
summarize the well-known used selection approaches as follows:
\begin{itemize}[leftmargin=*]
    \item{\textit{Cross-validation selection (CV)}:} select $\mathcal{M}^*_{K}$ using local validation accuracy;
    \item{\textit{Data selection (DS)}: $\mathcal{M}^*_{K} = \{M_i \, | \, i \in \mathtt{top}(\{n_1, \cdots, n_m\}, K)\}$}, i.e., 
                the models trained with the top $K$ size training dataset, which are selected by $\mathtt{top}$;
    \item{\textit{Random selection (RS)}:} $\mathcal{M}^*_{K}$ consists of model random selected from $\mathcal{M}$;
    \item{\textit{All selection (AS)}: select $\mathcal{M}$ as the target model set ignoring $K$, this method will consider all clients' data but will be very time-consuming.}
\end{itemize}

Besides, we construct the following baselines in terms of the model fusion, 
which derives a single model leading to the highest efficiency for inference,
as comparison with the traditional federated learning methods.
The final model $M^*$ is defined as:
\begin{itemize}
    \item{\textit{Federated averaging (FedAvg)}: $M^* = \sum_{i = 1}^{m} \frac{n_i}{\sum_{j=1}^{m}n_j} M_i$};
    \item{\textit{Mean averaging (MeanAvg)}: $M^* = \frac{1}{m}\sum_{i = 1}^{m} M_i$}.
\end{itemize}

Also, we include the following results as the ground-truths for comparison,
\begin{itemize}
    \item{\textit{Label distribution selection (LDS)}:} 
         utilizing the label distribution instead of model representation as the input of our method~\footnote{Note that the label distribution is unavailable in the real federated learning scenarios.};
    \item{\textit{Oracle}:} 
        using the aggregated dataset $ D = \bigcup_{i=1}^{m} D_i$ to 
    train a model $M_{oracle}$, whose performance is the `oracle'. 
\end{itemize}

\subsection{Performance Analysis}

    

\begin{figure}[t]
    \centering

    \subcaptionbox{homo \label{subfig:a1}}[.4\linewidth]
    {%
        \includegraphics[width =\linewidth]{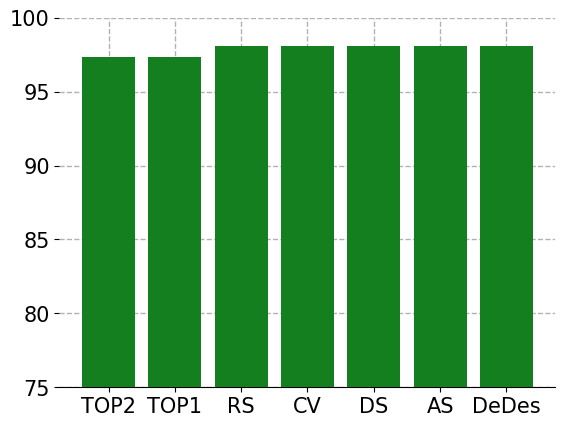}}
    \subcaptionbox{iid-dq \label{subfig:b1}}[.4\linewidth]
    {%
        \includegraphics[width =\linewidth]{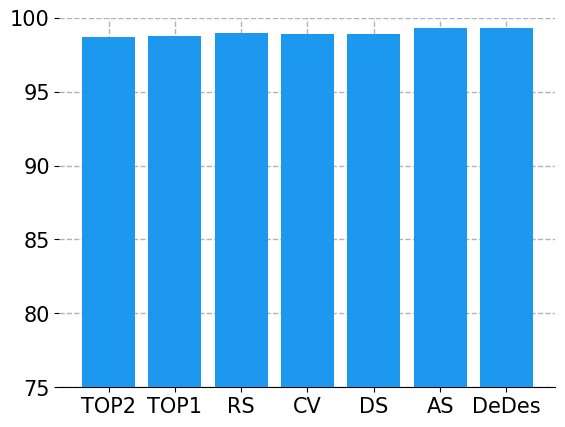}}\quad
    \subcaptionbox{noniid-lds \label{subfig:c1}}[.4\linewidth]
    {%
        \includegraphics[width =\linewidth]{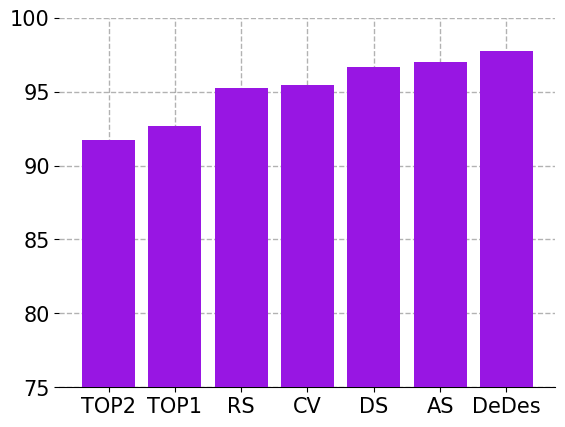}}
    \subcaptionbox{noniid-l$4$ \label{subfig:d1}}[.4\linewidth]
    {%
        \includegraphics[width =\linewidth]{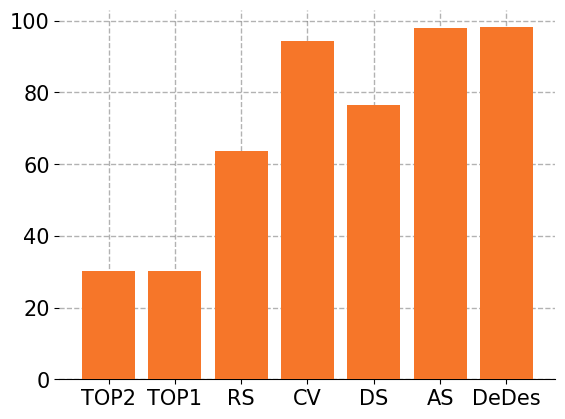}}
    \caption{Ensemble Learning (Weighted Voting) performance (test accuracy, \%) comparison on \textit{EMNIST Digits} dataset for $m$ = 200, $K$ = 80.}
    \label{fig:esp}
\end{figure}


\textbf{The effectiveness of ensemble learning} Figure \ref{fig:esp} compared different methods for 4 types of data partition settings, where \textit{TOP 1} and \textit{TOP 2} mean a single model who got the best and second best test accuracy on the whole test dataset $D^{test}$, i.e., $D^{test} = \bigcup_{i=1}^{m} D^{test}_i$, where $D^{test}_i$ is the test set for party/client $i$. As shown in Fig. \ref{fig:esp}, the performance of the ensemble methods (such as \textit{AS} and \texttt{DeDES\xspace}) are always better than single models, which validates the effectiveness of ensemble learning under one-shot federated learning settings. 


\textbf{Comparison of \methodName with other methods} For $m$ models, the number of possible ensemble teams is $2^m$, i.e., the number of possible ensemble teams increases exponentially with $m$. Since test all teams to get the optimal one is unpractical unless $m$ is very small, so in our experiment we will compare \methodName with existing methods to validate its superiority.

Table \ref{tab:performance} shows the test performance of selective configurations for different datasets and partition methods. As we can see, the performance of the \textit{Oracle} method is always the best, since it is the centralized setting and can get all parties' data/information; meanwhile, the performance of the \textit{FedAvg} or \textit{MeanAvg} is significantly worst (near random guess), with only test accuracy around 2\% for the \textit{EMNIST Balanced} datasets, which validates that directly average/fuse well-trained models are not suitable for the one-shot federated learning setting.

\begin{table*}[t]
    \centering
    \caption{Test accuracy (\%) comparison for different dataset on different data partitions and model structures. The best and next best methods are \textbf{bolded} and \underline{underlined}, respectively. If our \methodName method is better than the \textit{LD} ground-truth method, the value of \textit{LD} method will be marked in \textcolor[rgb]{ 0,  .439,  .753}{skyblue}. }
\resizebox{\columnwidth}{!}{%
\begin{tabular}{cccc|ccccccc|c|c}
\hline
Dataset & Partition & \textit{m} & \textit{K} & \methodName & AS & CV & DS & RS & FedAvg & MeanAvg & LD & Oracle \\
\hline
\multicolumn{1}{c}{\multirow{2}[8]{*}{\makecell{EMNIST Digits \\ (VGG-5 Spinal FC)}}} & homo  & 400   & 150   & 98.03 & \underline{98.10} & \textbf{98.10} & 98.08 & 98.07 & 10.28 & 10.26 & 98.10 & 99.74 \\
\cline{2-13}      & iid-dq & 400   & 150   & \textbf{99.27} & 98.75 & \underline{98.93} & 98.88 & 98.72 & 10.51 & 10.48 & 99.27 & 99.71 \\
\cline{2-13}      & noniid-ld & 400   & 150   & \textbf{97.67} & \underline{96.99} & 95.47 & 91.70 & 96.67 & 10.01 & 9.89  & \textcolor[rgb]{ 0,  .439,  .753}{92.86} & 99.72 \\
\cline{2-13}      & noniid-l3 & 400   & 150   & \textbf{98.21} & \underline{97.96} & 97.87 & 63.59 & 94.35 & 10.11 & 10.09 & \textcolor[rgb]{ 0,  .439,  .753}{98.13} & 99.61 \\
\hline
\multicolumn{1}{c}{\multirow{2}[8]{*}{\makecell{EMNIST Letters \\ (VGG-5 Spinal FC)}}} & homo  & 200   & 120   & 88.64 & 88.77 & \textbf{88.88} & \underline{88.82} & 88.68 & 3.72  & 3.71  & 88.77 & 95.12 \\
\cline{2-13}      & iid-dq & 200   & 120   & \underline{92.32} & 92.19 & 91.97 & \textbf{92.33} & 92.13 & 3.84  & 3.82  & 92.33 & 95.12 \\
\cline{2-13}      & noniid-ld & 200   & 120   & \textbf{87.93} & \underline{87.74} & 86.52 & 83.45 & 87.45 & 4.03  & 4.02  & \textcolor[rgb]{ 0,  .439,  .753}{85.01} & 94.90 \\
\cline{2-13}      & noniid-l8 & 200   & 120   & \textbf{89.10} & \underline{87.93} & 84.40 & 86.98 & 85.95 & 3.85  & 3.84  & \textcolor[rgb]{ 0,  .439,  .753}{87.54} & 95.06 \\
\hline
\multicolumn{1}{c}{\multirow{2}[8]{*}{\makecell{EMNIST Balanced \\ (VGG-5 Spinal FC)}}} & homo  & 100   & 50    & \textbf{85.19} & 84.94 & \underline{85.10} & 84.96 & 84.96 & 2.10  & 2.11  & \textcolor[rgb]{ 0,  .439,  .753}{84.83} & 89.70 \\
\cline{2-13}      & iid-dq & 100   & 50    & \underline{87.34} & 87.28 & 87.31 & \textbf{87.35} & 86.90 & 2.04  & 2.04  & 87.35 & 89.25 \\
\cline{2-13}      & noniid-ld & 100   & 50    & \textbf{83.43} & \underline{82.72} & 78.65 & 79.44 & 81.89 & 2.19  & 2.16  & \textcolor[rgb]{ 0,  .439,  .753}{77.28} & 89.48 \\
\cline{2-13}      & noniid-l18 & 100   & 50    & \textbf{85.43} & \underline{82.99} & 81.22 & 81.02 & 81.93 & 2.09  & 2.08  & \textcolor[rgb]{ 0,  .439,  .753}{82.87} & 89.52 \\
\hline
\multicolumn{1}{c}{\multirow{2}[8]{*}{\makecell{CIFAR10 \\ (Resnet-50)}}} & homo  & 200   & 100   & \underline{32.08} & \textbf{32.09} & 32.07 & 30.78 & 30.30 & 10.18 & 9.69  & 32.08 & 88.68 \\
\cline{2-13}      & iid-dq & 200   & 100   & 36.97 & 38.49 & \underline{38.84} & \textbf{39.03} & 36.66 & 10.04 & 10.03 & 38.81 & 88.10 \\
\cline{2-13}      & noniid-ld & 200   & 100   & \textbf{29.71} & \underline{29.23} & 26.02 & 29.10 & 26.67 & 9.89  & 9.88  & \textcolor[rgb]{ 0,  .439,  .753}{28.94} & 87.31 \\
\cline{2-13}      & noniid-l4 & 200   & 100   & \textbf{34.40} & \underline{33.50} & 32.24 & 30.00 & 33.05 & 10.02 & 9.87  & \textcolor[rgb]{ 0,  .439,  .753}{34.15} & 89.67 \\
\hline
\multicolumn{1}{c}{\multirow{2}[8]{*}{\makecell{CIFAR100 \\ (Resnet-50)}}} & homo  & 20    & 12    & \underline{20.84} & \textbf{22.84} & 20.58 & 20.65 & 20.48 & 0.99  & 0.99  & \textcolor[rgb]{ 0,  .439,  .753}{20.85} & 59.81 \\
\cline{2-13}      & iid-dq & 20    & 12    & \underline{47.38} & 47.37 & \underline{47.38} & \underline{47.38} & 25.10 & 1.00  & 0.94  & 47.38 & 60.35 \\
\cline{2-13}      & noniid-ld & 20    & 12    & \underline{16.31} & \textbf{18.71} & 15.97 & 16.15 & 15.78 & 0.96  & 0.97  & \textcolor[rgb]{ 0,  .439,  .753}{15.32} & 60.38 \\
\cline{2-13}      & noniid-l45 & 20    & 12    & \underline{21.29} & \textbf{23.68} & 20.56 & 20.26 & 19.97 & 0.92  & 0.91  & \textcolor[rgb]{ 0,  .439,  .753}{19.61} & 61.74 \\
\hline
\end{tabular}%
}
    \label{tab:performance}%
\end{table*}%

As demonstrated in Table \ref{tab:performance}, with the \textit{homo} partition, the accuracy difference between all methods is minimal, making it difficult to determine which method is superior. This is because the \textit{homo} partition is an IID setting, hence the data distribution of all parties is nearly identical. As a result, each party contains the same information as the others, so there is not a significant difference regardless of which parties we choose; for the \textit{iid-dq} partition, the \textit{Data Selection (DS)} is the best method for most of datasets, this is because under this setting, the single TOP 1/2 models as in Fig. \ref{fig:esp} (b) have the largest dataset with samples of every class in the label set $\{1, \ldots, C\}$, so the models themselves already have strong generalization ability. Therefore, under this partition, the more data we have, the better performance we will get, hence the best way is to select $K$ models with top $K$ largest data sets.

When the data partition is Non-IID (\textit{noniid-ld} and \textit{noniid-lk}), we can see that \methodName achieves the best performance for most of the datasets, with different $m$ and $K$ (more $m$ and $K$ combinations are in the $supplementary$), which validates the effectiveness of our method. \methodName can get the second best performance for the \textit{CIFAR100} dataset, with the \textit{AS} method be the best, this is because \textit{CIFAR100} has 100 labels, thus the data amount of each individual label for local parties is too tiny to train a generalized model. Under this condition, the $AS$ method will get more information than other methods and therefore have better performance. But for other datasets especially EMNIST where all local models are more generalized, \methodName will get better performance than others.

In some case \methodName is even better than the ground-truth label distribution selection (LDS), which validates that our model representation is very effective. 


\textbf{Complete Inspection on ensemble teams} When $m=10$, we can have $2^{10}=1024$ ensemble teams to select. Table \ref{tab:completeInspection} enumerated the accuracy of all 1024 teams and the ranking of selected teams generated by different approaches. We can see that the ensemble team selected by \methodName is ranked higher than other baseline methods, which validates the efficacy of our method.




\begin{table}
  \caption{Complete inspection on ensemble teams for \textit{EMNIST Balanced} dataset with $m=10, K=6$, \textit{noniid-lds} partition.}
  \label{tab:completeInspection}
  \begin{tabular}{c|c|c} \hline
    Method & Rank & Accuracy (\%) \\ \hline
    \methodName & \textbf{34/1024} & \textbf{84.34} \\ \hline
    AS    & 114/1024    & 83.39 \\ \hline
    CV    & 241/1024    & 82.29 \\ \hline
    DS    & 348/1024    & 80.86 \\ \hline
    LD    & 608/1024    & 77.77 \\ \hline
    RS    & 669/1024    & 76.54 \\ \hline
    \end{tabular}
\end{table}

\begin{figure}
  \includegraphics[width=0.65\linewidth]{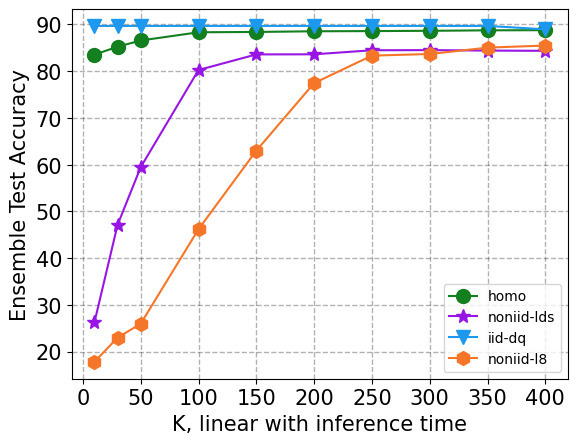}
  \caption{The relationship of $K$ and Ensemble Test Accuracy of \methodName for the \textit{EMNIST Letters} Dataset when $m$=400.}%
  \label{fig:impactK}
\end{figure}

\subsection{Impact on Efficiency}

Table. \ref{tab:performance} shows that in some cases, \methodName is the second best method after \textit{All Selection (AS)}. Note that the efficiency of \textit{AS} is quite poor, and the performance gap between these two approaches is small, validating that our method can reduce ensemble time to a large extent with minimal performance loss.


It is easy to know that the inference time for ensemble learning (weighted voting) increases linearly with $K$, i.e., the total inference time $T$ for one test sample is $K\times$$c$, where $c$ is constant inference time for one sample by one model. The experimental results depicted in Fig.\ref{fig:impactK} indicate that when $K$ reaches a certain value, the test accuracy will not increase significantly, sometimes even decrease. Therefore, with a suitable $K$ (usually 50\% of $m$), we can substantially reduce our inference time for ensemble learning while achieving good ensemble performance. And we do not need to concern too much about the running duration of \methodName compared to others because the ensemble selection process will only run once and will finish in a few minutes, therefore it is of little consequence.


    




\subsection{Ablation Studies}

For experiment details of this section, please refer to the supplementary.

\begin{itemize}
    \item \textbf{Performance Comparison on different model structures and datasets} Our method is solid for various model structures and datasets.
    \item \textbf{Performance Comparison on different model representation} It is better to use the models' later layer's parameters for representation than utilizing their front layer's parameters.
    \item \textbf{Importance of Dimension Reduction Methods.} \textit{Kernel-PCA} is better than other dimension reduction methods such as \textit{PCA} and \textit{non-compression}.
    \item \textbf{Clustering/Diversity validation} Our method can really cluster similar models together and the whole team's diversity is higher than other methods.
    

    
    
    
    

\end{itemize}








\section{\uppercase{Conclusion}}
\label{sec:conclusion}

This paper presents a novel \textit{Data-Free Diversity-Based} method called \methodName to address the ensemble selection problem for models generated by one-shot federated learning. Experiments demonstrated our method can achieve both better performance and efficiency for various model structures and datasets, especially for non-iid data partitions. To our knowledge, this is the first paper to systematically address the ensemble selection problem for one-shot federated learning, which is essential for applications such as machine learning model markets.

In the future, we will focus on the issue of heterogeneous model structures, propose more robust and useful model representation techniques, and better voting method to future improve ensemble performance and efficiency.

\bibliography{mybibliography} 

\begin{thebibliography}{10}

\bibitem{li2020federated}
Tian Li, Virginia Smith, et~al.
\newblock Federated learning: Challenges, methods, and future directions.
\newblock {\em IEEE Signal Processing Magazine}, 37(3):50--60, 2020.

\bibitem{geiping2020inverting}
Jonas Geiping, Hartmut Bauermeister, Hannah Droge, and Michael Moeller.
\newblock Inverting gradients-how easy is it to break privacy in federated
  learning?
\newblock {\em NeurIPS}, 33:16937--16947, 2020.

\bibitem{guha2019one}
Neel Guha, Ameet Talwalkar, and Virginia Smith.
\newblock One-shot federated learning.
\newblock {\em arXiv preprint arXiv:1902.11175}, 2019.

\bibitem{ijcai2021p205}
Qinbin Li, Bingsheng He, and Dawn Song.
\newblock Practical one-shot federated learning for cross-silo setting.
\newblock In Zhi-Hua Zhou, editor, {\em Proceedings of the Thirtieth
  International Joint Conference on Artificial Intelligence, {IJCAI-21}}, pages
  1484--1490. International Joint Conferences on Artificial Intelligence
  Organization, 8 2021.
\newblock Main Track.

\bibitem{vartak2016modeldb}
Manasi Vartak, Matei Zaharia, et~al.
\newblock Modeldb: a system for machine learning model management.
\newblock In {\em Proceedings of the Workshop on Human-In-the-Loop Data
  Analytics}, pages 1--3, 2016.

\bibitem{wolf2019huggingface}
Thomas Wolf, Morgan Funtowicz, et~al.
\newblock Huggingface's transformers: State-of-the-art natural language
  processing.
\newblock {\em arXiv preprint arXiv:1910.03771}, 2019.

\bibitem{sagi2018ensemble}
Omer Sagi and Lior Rokach.
\newblock Ensemble learning: A survey.
\newblock {\em Wiley Interdisciplinary Reviews: Data Mining and Knowledge
  Discovery}, 8(4):e1249, 2018.

\bibitem{zhou2002ensembling}
Zhi-Hua Zhou, Jianxin Wu, and Wei Tang.
\newblock Ensembling neural networks: many could be better than all.
\newblock {\em Artificial intelligence}, 137(1-2):239--263, 2002.

\bibitem{caruana2004ensemble}
Rich Caruana et~al.
\newblock Ensemble selection from libraries of models.
\newblock In {\em ICML}, 2004.

\bibitem{wu2021boosting}
Yanzhao Wu, Ling Liu, et~al.
\newblock Boosting ensemble accuracy by revisiting ensemble diversity metrics.
\newblock In {\em CVPR}, pages 16469--16477, 2021.

\bibitem{wu2021boosting2}
Yanzhao Wu and Ling Liu.
\newblock Boosting deep ensemble performance with hierarchical pruning.
\newblock In {\em ICDM}, pages 1433--1438. IEEE, 2021.

\bibitem{bonawitz2019towards}
Keith Bonawitz, Brendan McMahan, et~al.
\newblock Towards federated learning at scale: System design.
\newblock {\em Proceedings of Machine Learning and Systems}, 1:374--388, 2019.

\bibitem{duan2020self}
Moming Duan, Liang Liang, et~al.
\newblock Self-balancing federated learning with global imbalanced data in
  mobile systems.
\newblock {\em IEEE Transactions on Parallel and Distributed Systems},
  32(1):59--71, 2020.

\bibitem{kasturi2020fusion}
Anirudh Kasturi et~al.
\newblock Fusion learning: A one shot federated learning.
\newblock In {\em International Conference on Computational Science}, pages
  424--436. Springer, 2020.

\bibitem{bommasani2021opportunities}
Rishi Bommasani, Drew~A Hudson, Ehsan Adeli, Russ Altman, Simran Arora, Sydney
  von Arx, Michael~S Bernstein, Jeannette Bohg, Antoine Bosselut, Emma
  Brunskill, et~al.
\newblock On the opportunities and risks of foundation models.
\newblock {\em arXiv preprint arXiv:2108.07258}, 2021.

\bibitem{baylor2017tfx}
Denis Baylor, Levent Koc, et~al.
\newblock Tfx: A tensorflow-based production-scale machine learning platform.
\newblock In {\em Proceedings of the 23rd ACM SIGKDD International Conference
  on Knowledge Discovery and Data Mining}, pages 1387--1395, 2017.

\bibitem{zhou2021ensemble}
Zhi-Hua Zhou.
\newblock Ensemble learning.
\newblock In {\em Machine learning}. Springer, 2021.

\bibitem{schapire2013explaining}
Robert~E Schapire.
\newblock Explaining adaboost.
\newblock In {\em Empirical inference}, pages 37--52. Springer, 2013.

\bibitem{wang2019stacking}
Yuyan Wang et~al.
\newblock Stacking-based ensemble learning of decision trees for interpretable
  prostate cancer detection.
\newblock {\em Applied Soft Computing}, 77:188--204, 2019.

\bibitem{ma2015several}
Zhongchen Ma, Qun Dai, and Ningzhong Liu.
\newblock Several novel evaluation measures for rank-based ensemble pruning
  with applications to time series prediction.
\newblock {\em Expert systems with applications}, 42(1):280--292, 2015.

\bibitem{maskouni2018auto}
Mojtaba~Amiri Maskouni and Xiaofang Zhou.
\newblock Auto-ces: an automatic pruning method through clustering ensemble
  selection.
\newblock In {\em Australasian Database Conference}, pages 275--287. Springer,
  2018.

\bibitem{kuncheva2003measures}
Ludmila~I Kuncheva and Christopher~J Whitaker.
\newblock Measures of diversity in classifier ensembles and their relationship
  with the ensemble accuracy.
\newblock {\em Machine learning}, 51(2):181--207, 2003.

\bibitem{mchugh2012interrater}
Mary~L McHugh.
\newblock Interrater reliability: the kappa statistic.
\newblock {\em Biochemia medica}, 22(3):276--282, 2012.

\bibitem{yule1900vii}
Huaxiang Zhang and Linlin Cao.
\newblock A spectral clustering based ensemble pruning approach.
\newblock {\em Neurocomputing}, 139:289--297, 2014.

\bibitem{partridge1997software}
Derek Partridge and Wojtek Krzanowski.
\newblock Software diversity: practical statistics for its measurement and
  exploitation.
\newblock {\em Information and software technology}, 39(10):707--717, 1997.

\bibitem{zhao2019pyod}
Yue Zhao, Zain Nasrullah, and Zheng Li.
\newblock Pyod: A python toolbox for scalable outlier detection.
\newblock {\em JMLR}, 20(96):1--7, 2019.

\bibitem{li2022federated}
Qinbin Li, Yiqun Diao, Quan Chen, and Bingsheng He.
\newblock Federated learning on non-iid data silos: An experimental study.
\newblock In {\em ICDE}, pages 965--978. IEEE, 2022.

\end{thebibliography}
\bibliographystyle{unsrt} 

\newpage

\appendix

\begin{figure*}[t]
    \centering
    \includegraphics[width=0.8\textwidth]{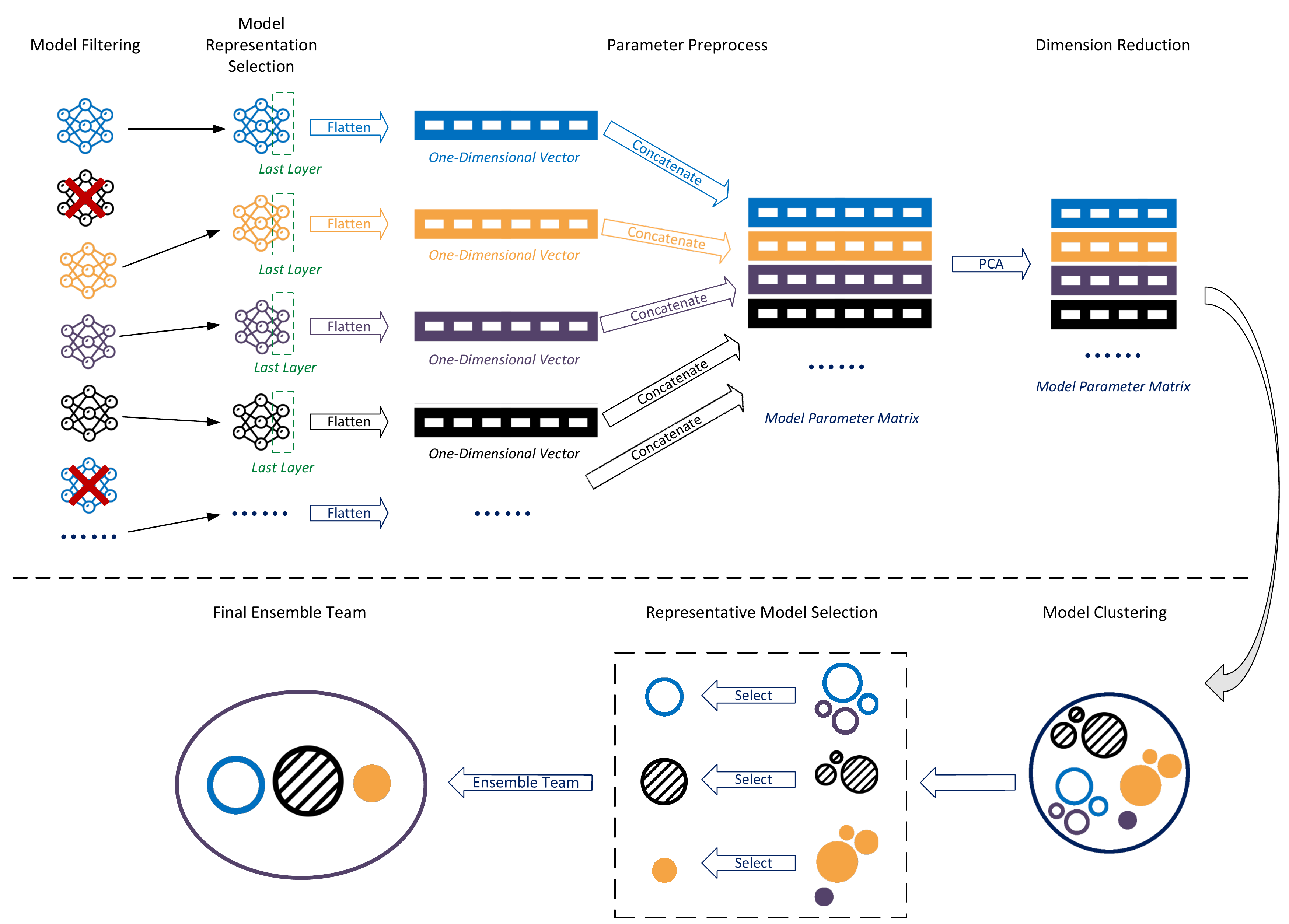}
    \caption{Flow chart of the execution process of \methodName framework, where last layer is used to represent the model and PCA is used as the dimension reduction method. Circles of the same style (same color and texture) represent models with actual high similarity.}
    \label{fig:est}
\end{figure*}
\subsection{Execution process of \methodName}
\label{sec:case}
Fig. \ref{fig:est} gives a flow chart of \methodName when we use  parameters of the model's last layer as the model representation and PCA as the dimension reduction method. The choice of clustering method will depend on the data partition which are shown in section 4.1.

\begin{figure}[t]
    \centering
    \includegraphics[width =.8\linewidth]{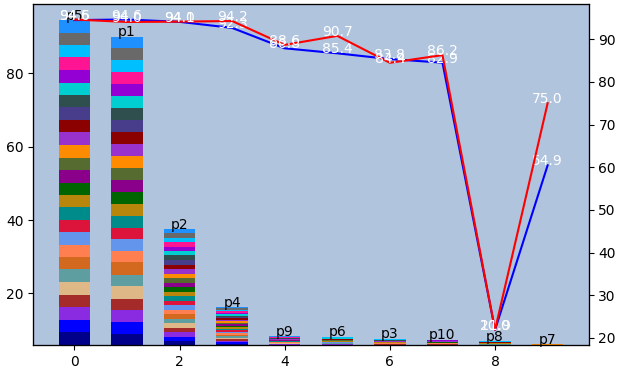}
    \caption{Data distribution for the iid-dq partition of the EMNIST letters dataset with local validation accuracy, and test accuracy for the whole test set $D^{test}$ when $m$=10. Every dot at the \textcolor{red}{red} line shows the best local validation accuracy after training for 200 epochs while the dots at the \textcolor{blue}{blue} line show the test accuracy (of the model with the best local validation accuracy) on the whole test set $D^{test}$ of the \textit{k}-th party p$k$.}
    \label{fig:outlier}
\end{figure}

\subsection{Model filter algorithm}

\begin{algorithm}[ht]
    \caption{ \texttt{OutlierFilter} algorithm for the model filtering}
    \label{alg:outlier_filer}
        \KwIn{model set $\mathcal{M}$, 
              truncated threshold pair ($p_{low}, p_{high}$), interval scale $s$,
              and model scores $\mathcal{S} = \{s_i\}_{i=1}^{m}$.}
        \KwOut{Outlier model set $\mathcal{O}$.}

        \nonl \textcolor{babyblueeyes}{\linecomment{Sort the score set $\mathcal{S}$, such as local validation accuracy, by ascending order.}}
        
        $\mathcal{S} \leftarrow AscendingSort(\mathcal{S})$

\nonl \textcolor{babyblueeyes}{\linecomment{Get the value of the $p_{low}$-th, $p_{high}$-th element of $\mathcal{S}$.}}
        
     $q_{low}, q_{high} \leftarrow \mathcal{S}{p_{low}}, \mathcal{S}{p_{high}}$
        
        $interval \leftarrow q_{high} - q_{low}$
        
        $outlier\_threshold \leftarrow q_{low} - s * interval$
        
        $\mathcal{O} \leftarrow \emptyset$
        
        \For{$i$ = 1 to $m$}{
            \If{$s_i < outlier\_threshold$}{
                $\mathcal{O} \leftarrow \mathcal{O} \cup \{M_i\}$
            }
        }
        \Return{$\mathcal{O}$}
\end{algorithm}

As in Alg.~1 from the \textit{main paper}, we use the $\mathtt{OutlierFilter}$ to obtain the outlier models $\mathcal{O}$ 
based on the model scores $\mathcal{S}$ provided from each party, 
which can be the local validation accuracy or prediction confidence.
$\mathtt{OutlierFilter}$ can be any score-based unsupervised outlier detection methods;
as we mentioned before,
we utilized a variation of the commonly-used box-plot in our experiment, which is shown in the above Alg.~\ref{alg:outlier_filer}.

As shown in Fig.~\ref{fig:outlier}, the 8-th party (i.e., p8) was not well-trained and not converged due to the inappropriate learning rate, which results in an inferior validation accuracy (20.9\%). Our method can successfully filter out this party's model by the model filtering method when selecting the ensemble team, and can then improve the final performance of ensemble learning.

\subsection{Experiment Setup}

\subsubsection{Component Configurations}
In this subsection, we will describe the default configurations of our \methodName framework for the experiments in the main paper.

For local model training, we utilize the \textit{SGD} optimizer to get 200 models for every party through 200 epochs of training with learning rate started at 0.1 and decrease at later epochs. I.e., we will save all the models from the 200 training rounds.  After the training finished, for every party, we select the model with the highest local validation accuracy (among these 200 models) as the final well-trained model and then upload it to the server of model market. Meanwhile, we will record the test results on the whole test set for this final model.

In our experiments, we select parameters of the final model layer (last layer) as the model representation; we utilize the MINMAX scaler to preprocess the \textit{Model Representation Matrix} (as shown in Fig. \ref{fig:est}), and the \textit{Gaussian Normalization} scaler to preprocess the \textit{Label Distribution} ground-truth input data; we do not utilize any dimension reduction strategy for the model representation matrix because the last layer of model parameters are already few in number. 

\textit{Spectral clustering} is used on the \textit{homo} and \textit{iid-dq} partitioned dataset; \textit{K-Means} clustering is used on the \textit{noniid-lds} and \textit{noniid-l$k$} partitioned dataset.

Our proposed model representative selection approach is utilized to determine the representative model within each cluster; as we have stated, we apply weighted voting strategy based on the size of local clients’ datasets to perform ensemble learning; we use the test accuracy on the whole test set $D^{test}$ as the evaluation metric for all the ensemble selection methods. 

\subsubsection{Environment}

All our experiments are running on a single machine with 1TB RAM and 256 cores AMD EPYC 7742 64-Core Processor @ 3.4GHz CPU. The GPU we used is NVIDIA A100 SXM4 with 40GB memory. 
The environment settings are: Python 3.9.12, PyTorch 1.12.1 with CUDA 11.6
on Ubuntu 20.04.4 LTS.

All the experimental results are the average over three trials.

\begin{figure*}[t]
    \centering
    \subcaptionbox{homo \label{subfig:a1}}[.23\linewidth]
    {%
        \includegraphics[width =\linewidth]{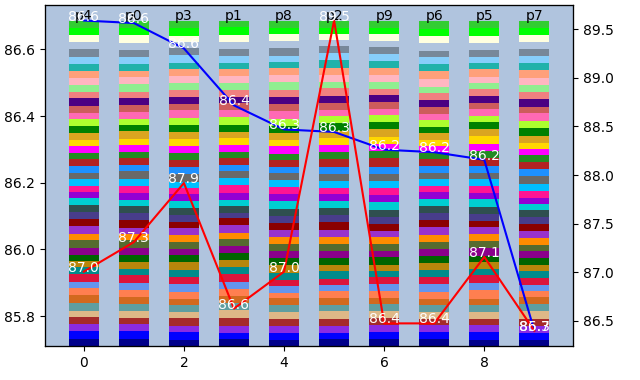}}
    \subcaptionbox{iid-dq \label{subfig:b1}}[.23\linewidth]
    {%
        \includegraphics[width =\linewidth]{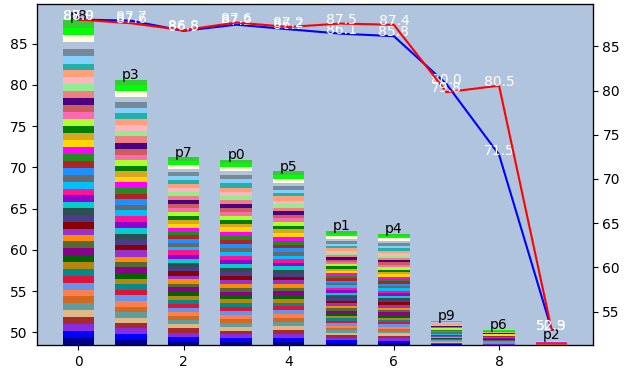}}\quad
    \subcaptionbox{noniid-lds \label{subfig:c1}}[.23\linewidth]
    {%
        \includegraphics[width =\linewidth]{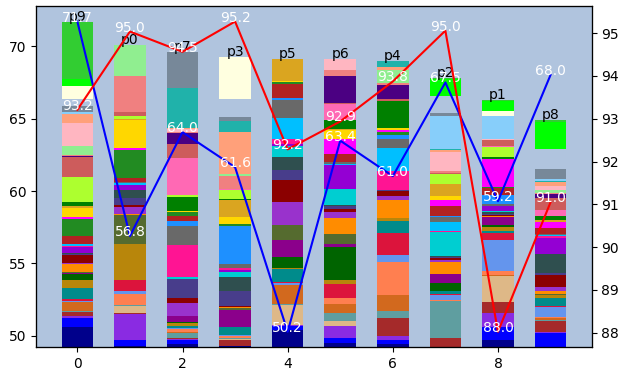}}
    \subcaptionbox{noniid-l$18$ \label{subfig:d1}}[.23\linewidth]
    {%
        \includegraphics[width =\linewidth]{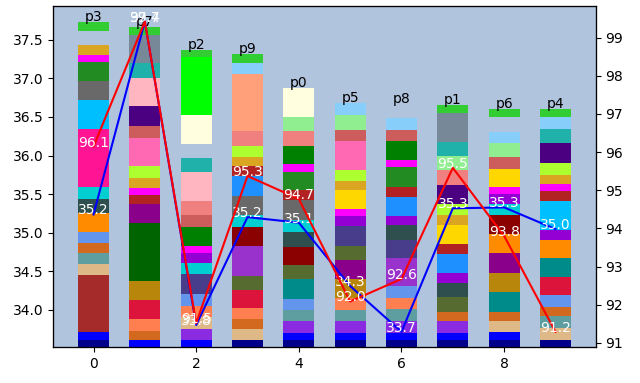}}
    
    \caption{Different data distribution of the EMNIST balanced dataset with local validation accuracy, and test accuracy for the whole test set when $m$=10. Every dot at the \textcolor{red}{red} line shows the best local validation accuracy after training for 200 epochs while the dots at \textcolor{blue}{blue} line show the test accuracy (of the model with the best local validation accuracy) on the whole test set $D^{test}$ of the \textit{k}-th party p$k$.}
    \label{fig:esp}
\end{figure*}

\subsection{Additional Experiments} 

In this section, we will show more experimental results as a supplement to the main paper. Due to the relatively large amount of experimental data, for each conclusion/finding, we take one of the experimental cases as presentation (one dataset, one $m$ and one $K$), while the conclusion remains similar to other datasets/$m$/$K$.

\subsubsection{The importance of ensemble learning}

Fig. \ref{fig:esp} shows the data distributions for the four data partitions of the EMNIST balanced dataset with local validation accuracy and test accuracy for the whole test set $D^{test}$ when $m$=10. We can see that the capacity (test accuracy for the whole test set) of a single model is weak, even when their local validation accuracies (validation accuracy for their own dataset) are high, which validates the importance of ensemble learning that can utilize the collaborative power of multiple models.

\begin{table*}[t]
    \centering
    \caption{Test accuracy (\%) comparison for different dataset on different data partitions and model structures. The best and next best methods are \textbf{bolded} and \underline{underlined}, respectively. If our \methodName method is better than the \textit{LD} ground-truth method, the value of \textit{LD} method will be marked in \textcolor[rgb]{ 0,  .439,  .753}{skyblue}. }
\resizebox{\columnwidth}{!}{%
\begin{tabular}{cccc|ccccccc|c|c}
\hline
Dataset & Partition & \textit{m} & \textit{K} & \methodName & AS & CV & DS & RS & FedAvg & MeanAvg & LD & Oracle \\
\hline
\multicolumn{1}{c}{\multirow{2}[8]{*}{\makecell{EMNIST Digits \\ (Resnet-50)}}} & homo  & 400   & 150   & 96.33 & \underline{96.46} & \textbf{96.65} & 96.23 & 96.31 & 10.25 & 10.22 & 96.70 & 99.71 \\
\cline{2-13}      & iid-dq & 400   & 150   & \textbf{98.13} & 98.01 & \underline{98.08} & 98.07 & 97.94 & 10.63 & 10.64 & 98.13 & 99.70 \\
\cline{2-13}      & noniid-ld & 400   & 150   & \textbf{96.52} & \underline{96.50} & 86.58 & 95.48 & 96.04 & 10.24 & 10.19 & \textcolor[rgb]{ 0,  .439,  .753}{94.80} & 99.70 \\
\cline{2-13}      & noniid-l3 & 400   & 150   & \underline{96.64} & \textbf{96.81} & 89.21 & 58.59 & 94.72 & 9.74  & 9.61  & 96.83 & 99.67 \\
\hline
\multicolumn{1}{c}{\multirow{2}[8]{*}{\makecell{EMNIST Letters \\ (Resnet-50)}}} & homo  & 200   & 120   & \underline{78.01} & 77.91 & \textbf{78.77} & 77.13 & 77.08 & 3.86  & 3.89  & \textcolor[rgb]{ 0,  .439,  .753}{77.41} & 94.76 \\
\cline{2-13}      & iid-dq & 200   & 120   & \underline{88.88} & \underline{88.88} & \underline{88.88} & \textbf{88.89} & 88.45 & 3.82  & 3.80  & \textcolor[rgb]{ 0,  .439,  .753}{88.85} & 95.13 \\
\cline{2-13}      & noniid-ld & 200   & 120   & \underline{79.78} & \textbf{80.55} & 79.53 & 77.27 & 78.65 & 4.23  & 4.28  & \textcolor[rgb]{ 0,  .439,  .753}{78.37} & 94.86 \\
\cline{2-13}      & noniid-l8 & 200   & 120   & \underline{81.10} & \textbf{82.79} & 80.54 & 78.86 & 80.28 & 3.74  & 3.69  & 82.33 & 95.08 \\
\hline
\multicolumn{1}{c}{\multirow{2}[8]{*}{\makecell{EMNIST Balanced \\ (Resnet-50)}}} & homo  & 100   & 50    & \underline{80.12} & 80.11 & \textbf{80.33} & 79.38 & 79.20 & 2.15  & 2.18  & \textcolor[rgb]{ 0,  .439,  .753}{78.93} & 89.44 \\
\cline{2-13}      & iid-dq & 100   & 50    & \underline{85.68} & \underline{85.68} & \underline{85.68} & \textbf{85.71} & 84.54 & 2.11  & 2.14  & 85.76 & 89.12 \\
\cline{2-13}      & noniid-ld & 100   & 50    & \underline{76.34} & \textbf{77.56} & 71.64 & 74.16 & 75.00 & 2.23  & 2.21  & \textcolor[rgb]{ 0,  .439,  .753}{70.75} & 89.52 \\
\cline{2-13}      & noniid-l18 & 100   & 50    & \underline{77.99} & \textbf{80.28} & 77.71 & 77.98 & 77.40 & 2.13  & 2.13  & 78.58 & 89.39 \\
\hline
\multicolumn{1}{c}{\multirow{2}[8]{*}{\makecell{CIFAR10 \\ (Densenet)}}} & homo  & 200   & 100   & \textbf{46.84} & 46.30 & \underline{46.47} & 45.37 & 45.68 & 10.49 & 10.08 & \textcolor[rgb]{ 0,  .439,  .753}{46.34} & 90.57 \\
\cline{2-13}      & iid-dq & 200   & 100   & 52.38 & 53.01 & \textbf{53.55} & \underline{53.47} & 51.76 & 10.45 & 10.56 & 54.03 & 90.38 \\
\cline{2-13}      & noniid-ld & 200   & 100   & \textbf{44.90} & \underline{43.91} & 40.89 & 40.54 & 41.49 & 10.38 & 10.52 & \textcolor[rgb]{ 0,  .439,  .753}{41.61} & 91.01 \\
\cline{2-13}      & noniid-l4 & 200   & 100   & \underline{47.04} & \textbf{48.51} & 46.08 & 40.92 & 45.43 & 9.71  & 9.47  & 47.45 & 90.79 \\
\hline
\multicolumn{1}{c}{\multirow{2}[8]{*}{\makecell{CIFAR100 \\ (Deep Layer Aggregation)}}} & homo  & 20    & 12    & \underline{23.01} & \textbf{24.80} & 22.47 & 22.27 & 22.63 & 0.95  & 0.94  & \textcolor[rgb]{ 0,  .439,  .753}{22.12} & 52.63 \\
\cline{2-13}      & iid-dq & 20    & 12    & \underline{39.18} & \underline{39.18} & \underline{39.18} & \underline{39.18} & 37.04 & 1.01  & 0.95  & 39.18 & 55.61 \\
\cline{2-13}      & noniid-ld & 20    & 12    & \underline{22.29} & \textbf{25.11} & 21.37 & 21.88 & 21.15 & 0.94  & 0.94  & \textcolor[rgb]{ 0,  .439,  .753}{21.42} & 55.94 \\
\cline{2-13}      & noniid-l45 & 20    & 12    & \underline{24.41} & \textbf{27.42} & 24.07 & 23.08 & 23.59 & 0.87  & 0.85  & \textcolor[rgb]{ 0,  .439,  .753}{23.19} & 54.90 \\
\hline
\end{tabular}%

}
    \label{tab:performance}%
\end{table*}%

\begin{table}
    \caption{Complete inspection on ensemble teams for \textit{EMNIST Digits} dataset with $m=10, K=5$, \textit{noniid-lds} partition.}
  \label{tab:completeInspection}
    \begin{tabular}{c|c|c} \hline
      Method & Rank & Accuracy (\%) \\ \hline
      \methodName & \textbf{214/1024} & \textbf{98.34} \\ \hline
      AS    & 372/1024    & 97.09 \\ \hline
      DS    & 608/1024    & 89.63 \\ \hline
      LD    & 675/1024    & 87.86 \\ \hline
      CV    & 933/1024    & 74.73 \\ \hline
      RS    & 952/1024    & 72.45 \\ \hline
      \end{tabular}
\end{table}

\begin{figure}[t]
  \includegraphics[width=0.65\linewidth]{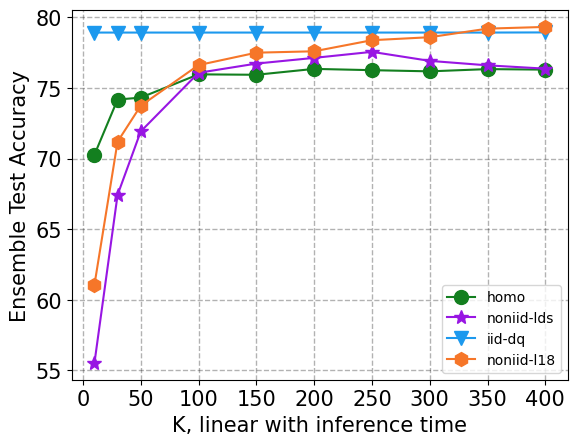}
  \caption{The relationship of $K$ and Ensemble Test Accuracy of \methodName for the \textit{EMNIST Balanced} Dataset when $m$=400.}%
  \label{fig:impactK}
\end{figure}


\subsubsection{Performance Analysis (Supplementary)}

Table. \ref{tab:performance} shows the performance of different methods when we apply them on the 5 datasets with 4 partitions, but different model structures than the main paper. $K$ is selected about half of $m$. Note that when $m=200, K=120$ on the EMNIST Letters dataset, the test accuracy of \texttt{DeDES} and AS, CV are the same, which means they both selected the same ensemble team. Compared to other baseline methods, \methodName can still achieve good performance (at least the second best, close to the best method of All Selection, which is very time-consuming).

Table \ref{tab:completeInspection} enumerated the accuracy of all 1024 teams and the ranking of ensemble teams selected by different methods for another data partition (\textit{noniid-lds}) of the EMNIST balanced dataset. We can see that the ensemble team selected by \methodName is ranked higher than other baseline methods, which validates the efficacy of our method. Note the value of $K$ here is 5 while the size of the best ensemble team among all 1024 teams is 4, therefore, how to select an appropriate $K$ remains an open problem.

\subsubsection{Impact on Efficiency (Supplementary)}

Fig.\ref{fig:impactK} gives another plot of the relationship between $K$ and test accuracy for the EMNIST Balanced dataset. The conclusion remains the same as the main paper that we don't have to select all models to form an ensemble team for most of the cases, which saves the inference time and also keeps good performance.


    




\begin{figure}[t]
  \includegraphics[width=0.7\linewidth]{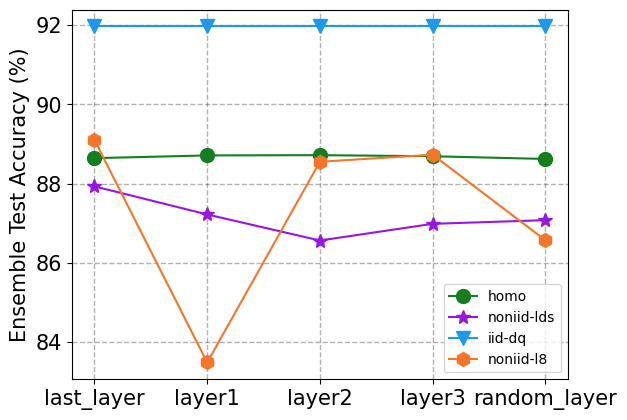}
  \caption{Comparison on the Ensemble Test Accuracy when applying different model representations on \methodName for the \textit{EMNIST Letters} Dataset, VGG-5 (Spinal FC) structure when $m$=200, $K=120$.}%
  \label{fig:layercomp}
\end{figure}

\begin{figure}[t]
  \includegraphics[width=0.7\linewidth]{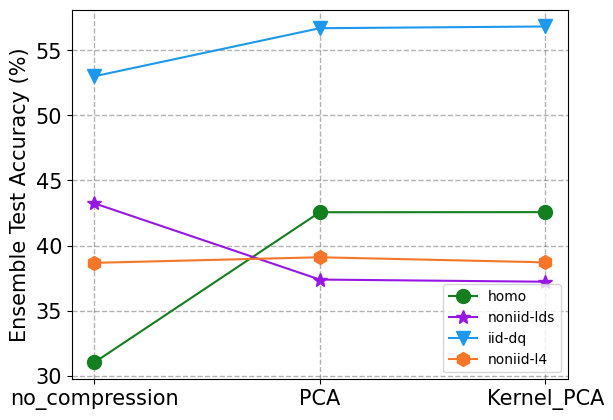}
  \caption{Comparison on the Ensemble Test Accuracy when applying different dimension reduction methods with the last layer as model representation on \methodName for the \textit{CIFAR10} Dataset, Resnet-50 structure when $m$=50, $K=30$.}%
  \label{fig:dimensionreduction}
\end{figure}

\subsubsection{Ablation Studies}

\begin{itemize}
    \item \textbf{Performance Comparison on different model structures and datasets} Our method is solid for various model structures and datasets. As shown in Table 2. in the main paper and Table. \ref{tab:performance} of this supplementary material, no matter what model structures/datasets we use, our method can achieve better performance than other baselines methods for ensemble learning.
    
    \item \textbf{Performance Comparison on different model representation} As shown in Fig. \ref{fig:layercomp}, for the VGG-5 (Spinal FC) model, layer1, layer2, layer3 and last\_layer represent the first, middle, latter and final/last fully-connected layers of the model which are selected as the model representations and the random\_layer means we randomly select 10\% of all layers as our model representation. As we can see, for the iid partitions (homo and iid-dq), almost no performance difference can be observed no matter what layer we choose; however, for the noniid partitions (noniid-lds and noniid-l18), there is still a gap in the performance of different layers as representations. From the figure we can see that it is better to use the models' later layer's parameters for representation than utilizing their front layer's parameters, but it is just a crude conclusion that doesn't apply to all situations. Therefore, how to select a good model representation to get better performance remains an open problem, especially for the noniid data partition.

    \item \textbf{Importance of Dimension Reduction Methods.} As shown in Fig. \ref{fig:dimensionreduction}, we compare three (dimension reduction) methods: PCA, Kernel\_PCA, and \textit{no\_compression} which means we don't compress the model representation (here is the parameters of model's last layer). For PCA and Kernel-PCA, we reduced the model representation to $m - sizeof(\mathcal{O})$ dimensions. We can see that for most of the partitions, the \textit{Kernel-PCA} is better than other methods such as \textit{PCA} and \textit{no\_compression}. This is because the Kernel-PCA can convert non-linear separable data to a new low-dimensional subspace suitable for alignment for linear classification, thus is suitable for the non-linear separable deep learning models. But we can also see that for the noniid-lds partition, we don't have to do the dimension reduction to get better ensemble learning results. Therefore, similar as the ablation study of model model representations, how to design a more effective dimension reduction method is still an open problem.

    \begin{figure}[t]
      \includegraphics[width=0.7\linewidth]{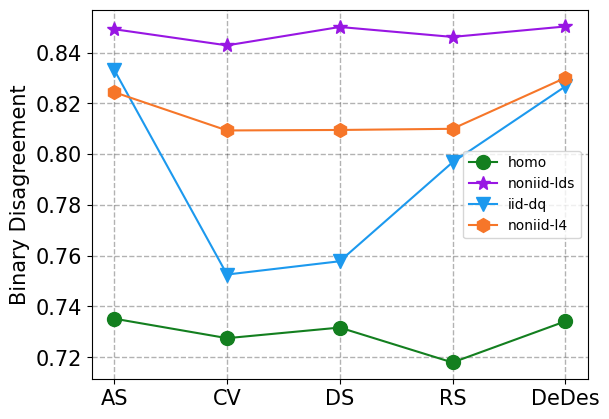}
      \caption{The Binary Disagreement value of ensemble teams selected by different methods for the \textit{CIFAR10} Dataset, Resnet-50 structure when $m$=50, $K=30$.}%
      \label{fig:bd}
    \end{figure}

    \begin{figure}[t]
      \includegraphics[width=0.7\linewidth]{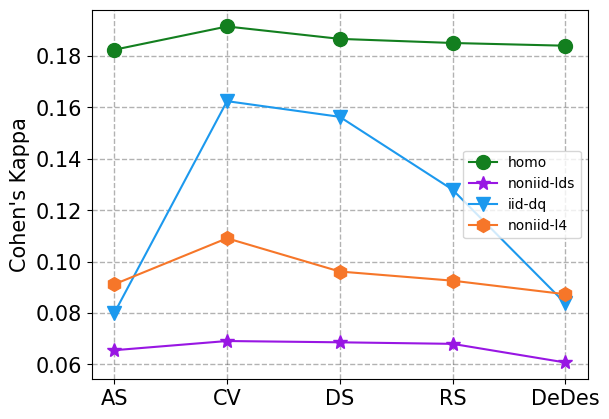}
      \caption{The Cohen's Kappa value of ensemble teams selected by different  methods for the \textit{CIFAR10} Dataset, Resnet-50 structure when $m$=50, $K=30$.}%
      \label{fig:cohen}
    \end{figure}
    
    \item \textbf{Clustering/Diversity validation} To validate our clustering results, we compare the Binary Disagreement (BD) \cite{kuncheva2003measures} and the Cohen's Kappa (CK) \cite{mchugh2012interrater} value of the ensemble teams selected by different methods to measure their diversities. The binary disagreement is defined as the ratio of the number of samples on which two models $M_i$ and $M_j$ get different prediction value to the total number of samples they predicted, higher binary disagreement means higher diversity; the cohen's kappa measures the agreement between two models in view of their reliability, lower cohen's kappa value indicates higher diversity (lower agreement). We take the average value of BD/CK for all pair ($M_i$, $M_j$) in $\modelset$ to get the final binary agreement/cohen's kappa value for the whole team $\modelset$.

    As shown in Fig. \ref{fig:bd} and Fig. \ref{fig:cohen}, compared to other baseline methods, the ensemble team's diversity of \methodName is higher (higher BD or lower CK), which also means the agreement of the whole team's models are low. Since we only select one model from every cluster, so this finding also indicates that our method can really cluster similar models together, which validates that \methodName can really generate an ensemble team with high diversity. Note that the All Selection (AS) method can also have high diversity compared to \methodName and meanwhile have high ensemble test accuracy, which validates the conclusion that the more diverse the models, the higher the ensemble's performance will have.
    

    
    
    
    

\end{itemize}

\end{document}